\pdfoutput=1
\documentclass[runningheads]{llncs}

\usepackage[T1]{fontenc}
\usepackage{amsmath}
\usepackage{amssymb}
\usepackage{graphicx}
\graphicspath{{figures/}}
\usepackage{booktabs}
\usepackage{algorithm}
\usepackage{algpseudocode}
\usepackage{cite}
\usepackage{url}
\usepackage[hidelinks]{hyperref}

\renewcommand{\orcidID}[1]{\unskip$^{\mbox{\href{https://orcid.org/#1}{[#1]}}}$}

\begin{document}

\title{Quality-Governed Deep Learning Network for Brain Tumor MRI Segmentation}
\titlerunning{Quality-Governed Deep Learning Network for Brain Tumor Segmentation}

\author{Asifiqbal~Saiyed\inst{1}\orcidID{0009-0006-1143-6530} \and
Nitesh~Chilakala\inst{2}\orcidID{0009-0000-3014-4564} \and
Jimmy~Joseph\inst{3}\orcidID{0009-0007-6947-7578} \and
Praneetha~Goli\inst{4} \and
Rohini~Maskar\inst{5}}

\authorrunning{A. Saiyed et al.}

\institute{Dept. of Math, Data \& Technology, Minot State University, Minot, ND, USA\\
\email{a.saiyeds@gmail.com}
\and
University of Houston Clear Lake, Dallas, TX, USA\\
\email{niteshcee72@gmail.com}
\and
Independent Researcher, Atlanta, GA, USA\\
\email{jimsweb@gmail.com}
\and
Capital One, Independent Researcher\\
\email{praneethagoli.712@gmail.com}
\and
Independent Researcher, San Jose, CA, USA\\
\email{rohinimaskar2@gmail.com}}

\maketitle

\begingroup
\renewcommand{\thefootnote}{}
\footnotetext{Accepted at the 7th International Conference on Computational
Vision and Bio Inspired Computing (ICCVBIC-2026).
\copyright~2026 IEEE. Personal use of this material is permitted. Permission
from IEEE must be obtained for all other uses, in any current or future media,
including reprinting/republishing this material for advertising or promotional
purposes, creating new collective works, for resale or redistribution to
servers or lists, or reuse of any copyrighted component of this work in other
works.}
\endgroup

\begin{abstract}
The segmentation of brain tumors from magnetic resonance imaging (MRI) is an
essential step for computer-aided neuro-oncology, treatment planning and
follow-up. While effective encoder-decoder representations are achieved by
U-Net and Attention U-Net, the skip sections of the network can generate
weak-contrast or noisy responses in the case of heterogeneous boundaries of the
tumor. In this paper, we introduce a Quality-Governed Deep Learning Network
(QGDL-Net) that can be used for binary tumor segmentation from FLAIR MRI
slices. The proposed model is able to predict the slice quality based on
contrast and SNR statistics and to fine-tune the encoder features prior to
decoding by learnable quality gates. We used the 80:20 train-validation split
for experiments with BraTS 2021. QGDL-Net achieved a Dice of 0.9747, IoU of
0.9506, precision of 0.9755, recall of 0.9739, F1 of 0.9747 and accuracy of
0.9982, outperforming U-Net and Attention U-Net. The EigenCAM maps and
segmentation overlays provide better tumor localization and interpretable
activation around the lesion regions.

\end{abstract}

\section{Introduction}
\label{sec:intro}

Accurate segmentation of glioma and other brain tumor regions from MRI supports
tumor volume estimation, radiation planning, surgical decision support and
longitudinal response assessment. In neuro-oncology, MRI is preferred due to
the complementary tissue contrast, pattern of edema and enhancing tumors with
FLAIR, T1, T1ce and T2 sequences. The BraTS project unified benchmark data and
assessment methods for this problem, allowing the algorithms to be compared in
a repeatable way~\cite{menze2015,bakas2017,baid2021}. But, the morphology of
the tumors is challenging due to different sizes, locations, intensities,
necrotic content and sharpness of the tumor margins.

Brain tumor segmentation has been revolutionised by deep learning, from
hand-crafted pipelines to trainable convolutional and transformer-based
systems. It has been shown that encoder-decoder models with skip connections
are good biomedical segmentation
baselines~\cite{ronneberger2015,cicek2016,isensee2021} with the aid of U-Net,
3D U-Net and nnU-Net. Subsequent works adopted attention gates, cascaded
networks, multiscale feature fusion and transformer modules, which further
enhanced the boundary localization and global context
modelling~\cite{oktay2018,wang2023transbts,hatamizadeh2022,jiang2022,zhou2023}.
All these advances, however, fail to account for the specific nature of the
response -- whether it is high-contrast tumor evidence or low-quality,
background-dominated responses -- in the current slice.

\begin{figure}[t]
\centering
\includegraphics[width=0.85\textwidth]{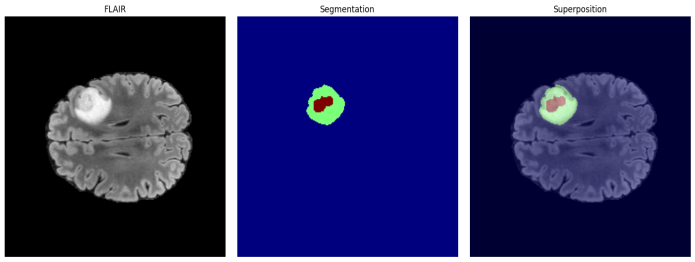}
\caption{Dataset view showing FLAIR input, tumor segmentation and
superposition.}
\label{fig:dataset}
\end{figure}

This work presents a quality-controlled and quality-governed segmentation
network called QGDL-Net, which evaluates the quality of the slices and adjusts
the features of the encoder according to their quality, before passing them on
to the decoder for fusion. According to the recent literature, feature quality,
receptive field design, attention and explainability are crucial for a
reliable medical
segmentation~\cite{zhang2024,zeineldin2024,preetha2025,shahid2025,tabassum2025}.
The main contributions are: (i) a quality descriptor based on contrast and
signal-to-noise statistics; (ii) channel-wise quality gates, placed inside skip
pathways as in U-Net; (iii) comparative tests against U-Net and Attention
U-Net; and (iv) visual validation by segmenting and overlaying on the basis of
the metrics and plotting a metric graph, using EigenCAM. Fig.~\ref{fig:dataset}
shows a representative BraTS2021 FLAIR MRI slice with its corresponding
segmentation mask and superimposed tumor region.

The motivation for the proposed method is twofold. First, MRI slices do not
equally contribute to the learning due to the presence of some slices having
large and distinct tumor regions, while others have weak borders, partial
appearance or the dominance of tissue in the background. Second, the standard
skip connections take low-level spatial information directly to the decoder,
and that can be helpful for retaining the structure, but noisy activations can
also be carried over. A quality-governed gate is a simple and trainable gate
that regulates this transfer.

The methodological comparison is done keeping all the pre-processing steps,
optimizer, and validation split the same for all the networks. The evaluation
employs both region-overlap metrics and visual evidence as segmentation models
are evaluated not just on a numerical basis, but also on the basis of the
anatomical plausibility of the mask predicted. The goal of practical
computer-aided diagnosis is to have an accurate, stable and explainable
segmentation model. The present work fulfills these requirements in three
complementary parts: (1) Segmentation backbone with added performance,
(2) Internal quality module and (3) EigenCAM visualization. The analysis shows
that QGDL-Net can boost the segmentation quality by selecting more informative
encoder features for reconstruction, confirming the claim.

The rest of this paper is divided into the following way.
Section~\ref{sec:lit} discusses some of the related literature.
Section~\ref{sec:method} proposes the system model and elaborates on the
offered QGDL-Net, and Section~\ref{sec:results} presents the analysis and
results of the experiments. Lastly, Section~\ref{sec:conclusion} is a
conclusion of the paper, including the future research directions.

\section{Literature Survey}
\label{sec:lit}

The segmentation benchmark and the stress on standardized evaluation in terms
of Dice and Hausdorff measures~\cite{menze2015} were highlighted. Bakas et al.\
also annotated the glioma images and assisted the evolution of BraTS
algorithms over the years~\cite{bakas2017}. Baid et al.\ enhanced the benchmark
in BraTS 2021 using a big multi-institutional cohort and a radiogenomic
classification task~\cite{baid2021}. These datasets could be used to perform
consistent algorithmic comparisons and show the presence of class imbalance
and heterogeneous tumor boundaries.

U-Net is a symmetric encoder-decoder network with skip connections that was
proposed in~\cite{ronneberger2015} for biomedical segmentation. \c{C}i\c{c}ek
et al.\ furthered this idea by creating volumetric segmentation with 3D
U-Net~\cite{cicek2016}. nnU-Net, proposed by Isensee et al., automatically made
many design decisions and served as a good reference method for biomedical
segmentation~\cite{isensee2021}. U-Net introduced attention gates to bypass
irrelevant areas, enabling the decoder to focus on relevant
structures~\cite{oktay2018}.

The current techniques are not limited to conventional convolutional
encoders. Transformer blocks were added to brain tumor segmentation by
TransBTS, exploiting global dependencies~\cite{wang2023transbts}. UNETR
proposed tokens for image patches for volumetric segmentation using a
transformer~\cite{hatamizadeh2022}, and SwinBTS and nnFormer investigated
hierarchical attention and the design of a volumetric
transformer~\cite{jiang2022,zhou2023}. Later published papers in the journal
introduced effective variants of U-Net enhanced by transformers, interpretable
transformer hybrids, multiscale attention U-Net, LIU-Net and meta-transformer
nnU-Net for brain tumor
MRI~\cite{zhang2024,zeineldin2024,preetha2025,shahid2025,tabassum2025}. While
these works inspire the proposed quality-governed design, QGDL-Net differs in
that it explicitly includes a quality score of the slices to modulate channels
of the encoder prior to skip fusion.

The first set of related works is that of convolutional encoder-decoder
segmentation. In the case of small medical structures, U-Net has developed the
concept of concatenating high-resolution features of the encoder with features
of the decoder, which is useful in this case. 3D U-Net expanded the concept to
volumetric learning, which is not only more memory-intensive but also demands
careful sampling of patches~\cite{cicek2016}. Systematic configuration of
normalization, spacing, patch size, loss and augmentation showed to be
sometimes as effective as picking a fixed architecture by
hand~\cite{isensee2021} in the case of nnU-Net.

The second group presents attention and transformer modules. U-Net suppresses
irrelevant skip features by means of attention using gating
coefficients~\cite{oktay2018}. Self-attention is capable of learning global
contexts which are hard for local
convolutions~\cite{wang2023transbts,hatamizadeh2022} by themselves, as is done
by TransBTS and UNETR. To alleviate the computation load of attention, SwinBTS
and nnFormer employ hierarchical windows or volumetric attention
structure~\cite{jiang2022,zhou2023}. These methods enhance representation
learning, but they typically do not obtain an explicit image quality score
prior to skip fusion.

The third class is models published recently in the journal that focus on
efficient, interpretable and transferable segmentation. ETUNet is based on the
transformer-enhanced U-Net design for 3D tumor segmentation~\cite{zhang2024}.
Explainable hybrid transformer frameworks report that interpretability can
enhance self-confidence in segmentation choices~\cite{zeineldin2024}. Recently,
feature fusion, lightweight design and pretrained segmentation knowledge are
key areas of ongoing research~\cite{preetha2025,shahid2025,tabassum2025}, with
multi-scale attention U-Net, LIU-Net and meta-transfer learning strategies
highlighting these. In this area, QGDL-Net is seen as a quality-controlled
feature-fusion approach.

Based on the literature reviewed, three gaps are found. First, networks with
high capacity can have high Dice scores, but can be very expensive to compute.
Second, attention mechanisms tend to learn about the importance of the spatial
location but may not explicitly learn the quality of the current slice at the
global level. Third, some methods claim numerical improvements, but do not
provide a concise mathematical description of the impact of the suggested
feature regulation on noise. The proposed QGDL-Net overcomes these limitations
with a lightweight quality estimator, channel-wise gates and a bounded-noise
argument.

\section{Proposed Methodology}
\label{sec:method}

\subsection{Dataset and Experimental Design}

The BraTS 2021 Task 1 training data was used in the study. Volumes and masks
corresponding to the different FLAIR were loaded from NIfTI files, as were
masks corresponding to the expert's segmentation. The volume segmented for
each patient was transformed to a binary mask by giving all labels of the
tumor to one class. The volumes were z-score normalized, those slices of the
axial slices with a tumor ratio $>$ a minimum threshold were selected, and the
images were resized to $128 \times 128$ pixels. The created slices were
divided into 80\% training and 20\% validation sets. The FLAIR sequence was
implemented in the prototype due to its ability to clearly show hyperintense
regions of many tumors and edema. While multimodal MRI can help improve
subregion segmentation, a single-modality setting is valuable to test whether
the proposed quality gate can aid in enhancing feature transfer within a
constrained input setting. The dataset characteristics and preprocessing
details used in this study are summarized in Table~\ref{tab:dataset}.

\begin{table}[t]
\centering
\caption{Dataset and preprocessing summary.}
\label{tab:dataset}
\begin{tabular}{ll}
\toprule
Item & Setting \\
\midrule
Dataset  & BraTS 2021 Task 1 \\
Input    & FLAIR MRI \\
Mask     & seg $> 0$ \\
Size     & $128 \times 128$ \\
Patients & 50 \\
Split    & 80:20 \\
Batch    & 16 \\
Platform & Kaggle GPU \\
\bottomrule
\end{tabular}
\end{table}

All compared networks underwent a similar pre-processing procedure.
Normalization of each volume was performed by dividing by the mean and
standard deviations of the respective volumes. Slices lacking adequate area of
tumour were discarded to prevent training of too many almost empty masks. The
same image-mask pairs are selected for U-Net, Attention U-Net and QGDL-Net to
avoid the impact of data imbalance and different sampling, so that the
differences in performance are due to the differences in structure. The task
was defined as a binary whole-tumor segmentation. BraTS labels contain many
tumor subregions, but in this study, all the labels of the tumor were
summarized to a single foreground class. For this purpose, this decision is
suitable to show the core quality-governed feature mechanism and isn't mixed
with multiclass loss weighing. The same gate could be expanded to improve
tumor segmentation, tumor core and whole tumor segmentation in the future.
Table~\ref{tab:models} presents the architectural configuration of the
compared segmentation models.

\begin{table}[t]
\centering
\caption{Compared models.}
\label{tab:models}
\begin{tabular}{lll}
\toprule
Model & Core mechanism & Role \\
\midrule
U-Net           & Skip encoder-decoder  & Baseline \\
Attention U-Net & Attention-gated skips & Attention baseline \\
QGDL-Net        & Quality-gated skips   & Proposed \\
\bottomrule
\end{tabular}
\end{table}

\subsection{Problem Formulation}

Let $x \in \mathbb{R}^{1 \times H \times W}$ be a normalized FLAIR slice, and
$y \in \{0,1\}^{H \times W}$ be the tumor mask. The objective is to learn
$f_\theta(x) = \hat{y}$, where $\hat{y} \in [0,1]^{H \times W}$ is a
pixel-wise tumor probability map. The final mask is obtained by applying a
threshold of 0.5 to $\hat{y}$.

The encoder progressively maps the input image into deeper feature spaces,
where spatial resolution decreases and semantic abstraction increases. The
decoder restores spatial resolution using transposed convolutions and feature
concatenation. This design allows coarse contextual information from deep
layers to be combined with fine anatomical details from early layers. The
limitation is that skip concatenation is unconditional, which motivates
quality-governed regulation.
\begin{equation}
F_l = \varphi\bigl(BN(W_l * F_{l-1} + b_l)\bigr)
\label{eq:enc}
\end{equation}
\begin{equation}
D_l = Conv\bigl([Up(D_{l+1}),\, F_l]\bigr)
\label{eq:dec}
\end{equation}

\subsection{Quality-Governed Feature Recalibration}

QGDL-Net computes a scalar quality descriptor for each slice using contrast
and signal-to-noise information:
\begin{equation}
c(x) = \operatorname{std}(x), \qquad
s(x) = \frac{\operatorname{mean}(x)}{\operatorname{std}(x) + \varepsilon}
\label{eq:cs}
\end{equation}
\begin{equation}
q(x) = 0.5\, c(x) + 0.5\, s(x)
\label{eq:q}
\end{equation}

For encoder feature $F_l$ with $C_l$ channels, a learnable gate maps $q$ into
channel weights:
\begin{equation}
w_l = \sigma\bigl(W_{2l}\, \delta(W_{1l}\, q + b_{1l}) + b_{2l}\bigr)
\label{eq:gate}
\end{equation}
\begin{equation}
F_l^{q} = w_l \otimes F_l
\label{eq:recal}
\end{equation}

Here $\sigma$ is the sigmoid function, $\delta$ is ReLU and $\otimes$ denotes
channel-wise multiplication. The decoder receives the quality-recalibrated
features $F_1^q$, $F_2^q$ and $F_3^q$. Since $0 \le w_l \le 1$, the gate
cannot amplify noisy encoder energy. If $F = S + N$, where $S$ denotes lesion
signal and $N$ denotes disturbance, then
$\|w_l N\|_2^2 = w_l^2 \|N\|_2^2 \le \|N\|_2^2$. Thus, quality gating bounds
non-informative feature energy while the loss encourages channels useful for
tumor prediction to keep larger weights.

\begin{figure}[t]
\centering
\includegraphics[width=0.85\textwidth]{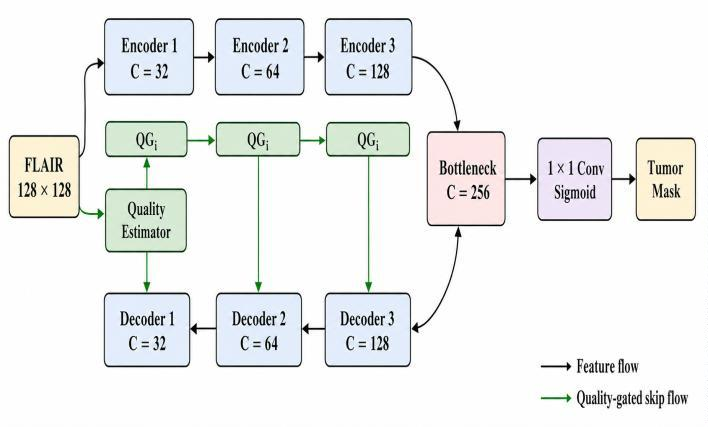}
\caption{Proposed QGDL-Net architecture with quality estimator and
quality-gated skip connections.}
\label{fig:architecture}
\end{figure}

During the training phase, the parameters of the gates are adjusted by
gradients of the Dice and BCE losses to better preserve channels that are
associated with the tumor overlap, while the influence of channels that are
repeatedly assigned false positives is reduced. This mechanism is easier to
compute as only a scalar quality input and two fully connected layers are used
at each level of the gate. Quality Gates are different than attention gates.
Attention U-Net emphasizes the spatial relevance of the features in the
decoder and features in the encoder, and QGDL-Net recalibrates the channels
based on the overall quality of the slice. Thus, the quality mechanism can be
viewed as reliability control of features, while the spatial attention
mechanism can be viewed as a selection of locations. To maintain the compact
architecture, the proposed model is based on the former. The overall
architecture of the proposed QGDL-Net is illustrated in
Fig.~\ref{fig:architecture}.

\begin{algorithm}[t]
\caption{Quality-Governed Training and Inference}
\label{alg:qgdl}
\begin{algorithmic}[1]
\Require MRI slices $X$, masks $Y$, models
$M = \{$U-Net, Attention U-Net, QGDL-Net$\}$, epochs $E$
\Ensure Trained segmentation model and probability mask $\hat{Y}$
\State Normalize FLAIR volume and extract tumor-containing axial slices.
\For{epoch $e = 1$ to $E$}
  \For{each mini-batch $(x, y)$}
    \State Compute $c(x)$, $s(x)$ and quality score $q(x)$.
    \State Extract encoder maps $F_1, F_2, F_3$.
    \State Estimate $w_1, w_2, w_3$ using quality gates.
    \State Recalibrate $F_l^q = w_l \otimes F_l$ and decode $\hat{Y}$.
    \State Update $\theta$ by minimizing $0.5\mathcal{L}_\beta + 0.5\mathcal{L}_d$.
  \EndFor
\EndFor
\State \Return binary mask $\hat{Y} \ge 0.5$ during inference.
\end{algorithmic}
\end{algorithm}

\subsection{Loss Function and Evaluation Metrics}

All models were optimized using a compound loss:
\begin{equation}
\mathcal{L} = 0.5\, \mathcal{L}_\beta + 0.5\, \mathcal{L}_d
\label{eq:loss}
\end{equation}
\begin{equation}
\mathcal{L}_d = 1 - \frac{2 \sum_i y_i \hat{y}_i + \varepsilon}
{\sum_i y_i + \sum_i \hat{y}_i + \varepsilon}
\label{eq:dice}
\end{equation}
where $\mathcal{L}_\beta$ denotes the binary cross-entropy (BCE) loss and
$\mathcal{L}_d$ denotes the Dice loss defined in~\eqref{eq:dice}.

Performance was evaluated using Dice, IoU, precision, recall, F1-score,
accuracy, ROC analysis and confusion matrices. EigenCAM was used to visualize
activation regions of the proposed network. The training configuration and
experimental settings are listed in Table~\ref{tab:training}.

\begin{table}[t]
\centering
\caption{Training configuration.}
\label{tab:training}
\begin{tabular}{ll}
\toprule
Parameter & Value \\
\midrule
Optimizer & Adam \\
Loss      & 0.5 BCE + 0.5 Dice \\
LR        & 0.001 \\
Threshold & 0.5 \\
Epochs    & 10/25/50 \\
Criterion & Best Dice \\
\bottomrule
\end{tabular}
\end{table}

\subsection{Implementation and Reproducibility Details}

The implementation was created with PyTorch and run in a Kaggle GPUs
environment. To ensure repeatability, the same dataset object, data split,
mini-batch size and loss function were used in three models. According to the
validation Dice, validation model checkpoints were saved separately. To let the
gate and attention parameters remain stable, the Attention U-Net and QGDL-Net
were trained for longer epochs while the baseline U-Net was evaluated for fewer
epochs, as it converged rapidly. The chosen evaluation protocol was carefully
chosen to be simple and clear. Each validation mask was segmented into
foreground and background pixels to calculate precision, recall, F1-score,
accuracy and IoU. The Dice measure was calculated using the intersection of
the predicted binary mask and the ground truth mask. Before thresholding, the
raw probability maps used to compute ROC curves were used to show how well the
pixels from the tumor and background could be separated across various
thresholds.

While the model is learnt in slices of 2D images, it retains the essence of
learning features guided by quality. The same quality gate can be generalized
to 3D feature tensors by using 3D convolutions and calculating $q$ from a
volume/patch. In this work, the adopted score for quality is deliberately
light. It integrates the contrast with signal-to-noise properties, instead of
using an extra neural quality evaluation network. This will not result in the
proposed method becoming significantly bigger than the baselines. Furthermore,
it simplifies the mechanism to interpret: high-quality slices have stronger
feature retention, while weak or noisy slices are controlled prior to decoder
fusion.

\section{Results and Discussion}
\label{sec:results}

Table~\ref{tab:performance} shows the quantitative results. The highest Dice,
IoU, precision, recall, F1-score and accuracy were achieved by QGDL-Net. The
Dice score of U-Net rose from 0.9471 to 0.9747 for QGDL-Net. The IoU for the
model rose from 0.8995 to 0.9506, which means that the model has a higher
degree of overlap with the expert mask and has fewer errors along the
boundaries. QGDL-Net was further advanced than the U-Net with the
quality-guided feature selection factor, while the attention mechanism also
served to enhance performance over the U-Net. The relative improvement of
QGDL-Net over the baseline models is reported in Table~\ref{tab:improvement}.

\begin{table}[t]
\centering
\caption{Performance comparison.}
\label{tab:performance}
\begin{tabular}{lcccccc}
\toprule
Model & Dice & IoU & Prec. & Rec. & F1 & Acc. \\
\midrule
U-Net           & 0.9471 & 0.8995 & 0.9507 & 0.9435 & 0.9471 & 0.9963 \\
Attention U-Net & 0.9629 & 0.9285 & 0.9664 & 0.9594 & 0.9629 & 0.9974 \\
QGDL-Net        & \textbf{0.9747} & \textbf{0.9506} & \textbf{0.9755} &
                  \textbf{0.9739} & \textbf{0.9747} & \textbf{0.9982} \\
\bottomrule
\end{tabular}
\end{table}

\begin{table}[t]
\centering
\caption{Improvement summary.}
\label{tab:improvement}
\begin{tabular}{llll}
\toprule
Comparison & Dice gain & IoU gain & Interpretation \\
\midrule
QGDL vs U-Net     & +0.0276 & +0.0511 & Better overlap \\
QGDL vs Att-U-Net & +0.0118 & +0.0221 & Added quality value \\
Visual            & Cleaner & Focused & Less leakage \\
\bottomrule
\end{tabular}
\end{table}

The most significant gain over U-Net is achieved in IoU with an improvement of
+0.0511 for QGDL-Net. This gain implies that the quality gate benefits the
spatial conformity of the entire forecasted area, as IoU penalizes missing
tumor pixels as well as extra predicted foreground. The precision value of
0.9755 also shows that QGDL-Net decreases background leakage from the
baselines.

Although this is less significant than the improvement of Attention U-Net, it
is still noticeable. There is no great concern for a significant performance
gap since spatial gating is already used by Attention U-Net. The results of
the additional measures, Dice gain (+0.0118) and IoU gain (+0.0221), however,
indicate that channel regulation using quality information also contributes to
the spatial attention. This confirms the hypothesis of a difference between
the reliability of features and the relevance of their spatiality in
segmentation.

The improvement in recall from 0.9435 to 0.9739 indicates that the QGDL-Net
performs better in terms of recall compared to the U-Net. The reason for this
is that under-segmentation may result in a reduced estimated tumor volume. The
precision gain demonstrates that the proposed method is able to control
false-positive regions, which is important when tumor regions are adjacent to
normal hyperintense tissues. The Dice coefficient and the F1 score of QGDL-Net
are equal because they both rely on the same foreground overlap formulation
and are derived for binary segmentation. For all images, the accuracy values
are high since most of the image pixels are background. So, just accuracy
isn't enough to prove that it's the best; it's the combined improvement in
Dice, IoU, precision and recall.

The performance of QGDL-Net is 2.91\% and 5.68\% better than U-Net in terms of
Dice score and IoU, respectively. The proposed model achieves a Dice metric
improvement of 1.23\% and an IoU metric improvement of 2.38\% against
Attention U-Net. While these differences are numerically small, they can make
a significant visual difference in the boundary accuracy in challenging slices
when measuring in segmentation.

\begin{figure}[t]
\centering
\includegraphics[width=\textwidth]{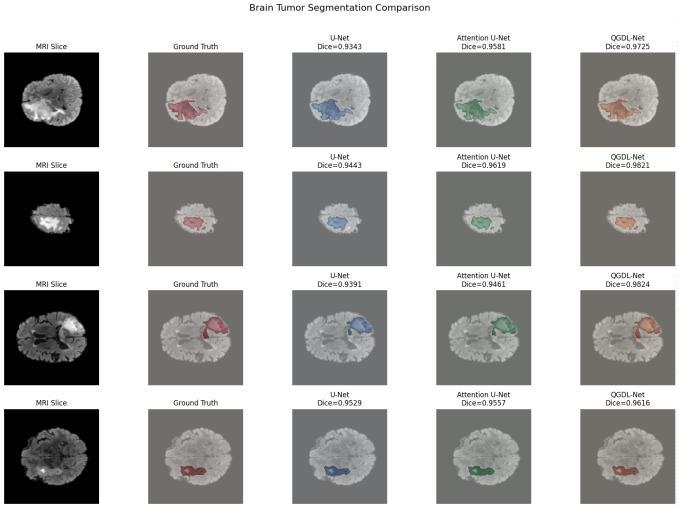}
\caption{Segmentation comparison for representative validation slices.}
\label{fig:segmentation}
\end{figure}

The segmentation masks obtained in the proposed method are more compact and
better aligned with the ground truth as shown in the segmentation overlays of
Fig.~\ref{fig:segmentation}. This is especially noticeable for small and
irregular tumor areas with only one baseline U-Net, where over-segmentation and
under-segmentation occur. The improvement is in line with the way the quality
gate is supposed to work: it prevents weak background activations from
reaching the skip fusion stage.

These visual differences are important for clinical decision-making, for
instance, to calculate tumor burden from segmentation maps or for directing
subsequent radiomic feature extraction. A mask that extends into normal tissue
can affect volume and texture features, and a mask that is not completely
around the tumor can lead to underestimation of tumor dimensions. The proposed
model minimizes both errors for the representative samples. The visual
comparison also shows an important practical benefit of the quality gating.
Tumor texture is bright but irregular in slices where texture is present;
baseline predictions could fragment or spill over into neighbouring tissue.
QGDL-Net creates smoother and more anatomically consistent masks as it is a
decoder that gets recalibrated features, not raw encoder features. This
enhances visual coherence without the need for post-processing.

QGDL-Net only introduces small fully connected Quality Gates as an additional
layer to the baseline encoder from a computational standpoint. Major
evolutionary changes to the convolutional pathway are still U-Net-style, which
makes the proposed model easier to train than large transformer models.

\begin{figure}[t]
\centering
\includegraphics[width=0.6\textwidth]{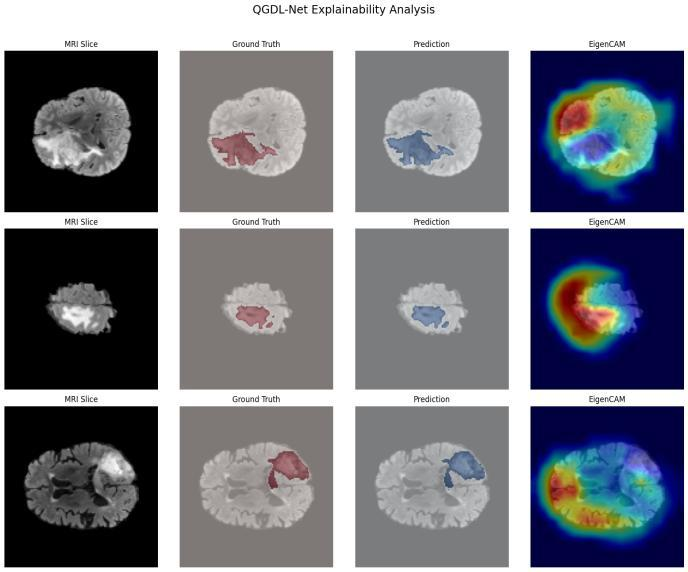}
\caption{EigenCAM explainability map of QGDL-Net.}
\label{fig:eigencam}
\end{figure}

\begin{figure}[t]
\centering
\includegraphics[width=0.7\textwidth]{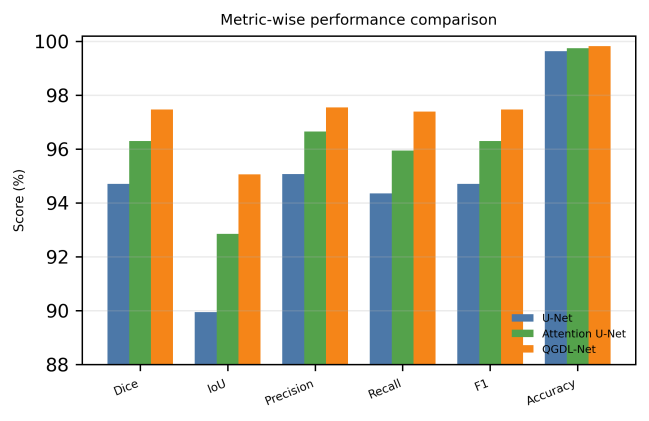}
\caption{Graph plot comparing Dice, IoU, precision, recall, F1 and accuracy.}
\label{fig:metrics}
\end{figure}

High-response areas are concentrated around the tumor-bearing areas, as seen
in Fig.~\ref{fig:eigencam} using EigenCAM visualization. EigenCAM is not a
validation process, but it offers a qualitative assessment that QGDL-Net
utilizes lesion-relevant regions. The metric plot (Fig.~\ref{fig:metrics})
also indicates that the performance does not improve solely with one metric,
but it is seen to improve across all the metrics that measure overlap and
pixel-classification.

The same can be seen from the confusion matrix and the ROC curve of the
experiment too. The predicted probability maps result in good
foreground-background separation, and the confusion matrices reveal fewer FP
and FN errors for the proposed model. These results match the numerical
results on precision and recall improvement.

One limitation of this study is that only binary FLAIR-based segmentation was
assessed. A multimodal and multiclass BraTS segmentation would give a more
robust clinical experiment. Furthermore, the current validation split is an
internal one, so an external test should be conducted between the institution
and scanners prior to deployment. The controlled comparison proves that
quality-guided gating can help improve a U-Net-style architecture even with a
smaller transformer backbone.

\subsection{Comparative Interpretation}

A performance improvement is observed when comparing the performance of U-Net,
Attention U-Net and QGDL-Net. The encoder-decoder baseline is given by U-Net.
This is enhanced by learning spatial relevance for skip connections, which is
done by an attention U-Net. QGDL-Net further enhances the performance by
making image quality dependent on feature transfer. Thus, there is a logical
experimental progression, not a random choice of baseline models, in the
three-model comparison.

QGDL-Net is not a substitute for attention or transformer research but an
orthogonal idea. Future systems could merge spatial attention with the
transformer context and quality-governed gates into a single network. In the
present work, the quality component is separated so its effect is clear.
Isolation is good for methodological papers because it minimises the problem
of attributing improvement to more than one unchecked change.

\subsection{Practical Implications}

For radiology support tools, robustness is as crucial as peak accuracy. A
model which is effective for high-contrast slices will not work in subtle
tumor areas. Through feature transfer based on quality statistics, QGDL-Net
prompts the network to adapt to the reliability level at the slice level.
This is especially helpful in data sets obtained from various scanners,
protocols and institutions.

The explainability result also lends credence to the interpretability.
EigenCAM maps are focused around tumor regions and overlap with the general
location of predicted maps. This isn't a clinical correctness check, but
rather a way to check if the network is using plausible image regions. This
visualization, together with overlays on the segments and quantitative
measurements, can increase confidence in the model output.

\section{Conclusion}
\label{sec:conclusion}

This paper introduced QGDL-Net for brain tumor magnetic resonance imaging (MRI)
segmentation in the context of quality control. The proposed method performs
encoder feature recalibration prior to decoder fusion with the help of a
slice-level quality descriptor and learnable channel gates. Using mathematical
analysis, it was found that sigmoid quality gates were unable to boost the
energy of noisy features, and experiments conducted on BraTS 2021 yielded
better performance than U-Net and Attention U-Net. QGDL-Net achieved a Dice of
0.9747 and an IoU of 0.9506. The method will be expanded to multimodal 3D
segmentation of subregions, external validation and uncertainty-aware clinical
deployment (future work).

The findings indicate that there is still a role for compact reliability-aware
modules in today's age of large transformer and hybrid networks. QGDL-Net does
not add depth but enhances the usefulness of the information already available
in the encoder features. The method is appealing for resource-poor experiments
and for applications where simplicity of computation and interpretability are
crucial.

Future direction is uncertainty estimation. A clinically useful segmentation
system ought to not only generate a mask but also convey confidence. A
combination of Quality gates with Monte Carlo dropout and deep
ensembles/calibration losses can be used to detect uncertain borders. These
additions would transform the network to be more radiotherapy planning and
radiomic-friendly.

Another implementation extension is to use a small quality vector instead of
a scalar quality value, consisting of contrast, entropy, edge strength,
tumor-area prior and local intensity variation. Such a vector would enable the
gate to discriminate between various causes of poor segmentation, such as low
contrast, small lesion size and noisy background tissue. In clinical
translation, the predicted mask is to be evaluated along with the
radiologist's review. QGDL-Net can be used as a first approximation for
contour generation and then corrected by hand. This would save time for
annotating the data, but at the same time provide expert supervision, an
essential condition for safety-critical applications like neuro-oncology.


\end{document}